\documentclass[a4paper,twoside]{article}
\usepackage{epsfig}
\usepackage{calc}
\usepackage{amssymb}
\usepackage{amstext}
\usepackage{amsmath}
\usepackage{natbib}
\usepackage{dsfont}
\usepackage{amsthm}
\usepackage{multicol}
\usepackage{pslatex}
\usepackage{apalike}
\usepackage{algorithm2e}

\usepackage{booktabs}
\usepackage[table,xcdraw]{xcolor}
\usepackage{array}
\usepackage{graphicx}         
\usepackage{subcaption}       
\usepackage{pgfplots}        
\usepackage{pgfplotstable}   
\usepackage{float}
\usepackage{url}
\usepackage[colorlinks=true, allcolors=blue]{hyperref}
\usetikzlibrary{plotmarks}

\usepackage{pgfplotstable}
\usepgfplotslibrary{fillbetween}

\usepackage[bottom]{footmisc}
\usepackage{SCITEPRESS}     

\definecolor{BL}{HTML}{003A7D}
\definecolor{LB}{HTML}{008DFF}
\definecolor{PI}{HTML}{FF73B6}
\definecolor{PU}{HTML}{C701FF}
\definecolor{GR}{HTML}{4ECB8D}
\definecolor{OR}{HTML}{FF9D3A}
\definecolor{RE}{HTML}{D83034}

\begin{document}

\title{Upside-Down Reinforcement Learning for More Interpretable Optimal Control}

\author{\authorname{Juan Cardenas-Cartagena\sup{1}, Massimiliano Falzari\sup{1}, Marco Zullich\sup{1} and Matthia Sabatelli\sup{1}}
\affiliation{\sup{1}Bernoulli Institute, University of Groningen, Groningen, The Netherlands}
\email{m.sabatelli@rug.nl}
}

\keywords{Upside-Down Reinforcement Learning, Neural Networks, Random Forests, Interpretability, Explainable AI}

\abstract{Model-Free Reinforcement Learning (RL) algorithms either learn how to map states to expected rewards or search for policies that can maximize a certain performance function. Model-Based algorithms instead, aim to learn an approximation of the underlying model of the RL environment and then use it in combination with planning algorithms. Upside-Down Reinforcement Learning (UDRL) is a novel learning paradigm that aims to learn how to predict actions from states and desired commands. This task is formulated as a Supervised Learning problem and has successfully been tackled by Neural Networks (NNs). In this paper, we investigate whether function approximation algorithms other than NNs can also be used within a UDRL framework. Our experiments, performed over several popular optimal control benchmarks, show that tree-based methods like Random Forests and Extremely Randomized Trees can perform just as well as NNs with the significant benefit of resulting in policies that are inherently more interpretable than NNs, therefore paving the way for more transparent, safe, and robust RL.}

\onecolumn \maketitle \normalsize \setcounter{footnote}{0} \vfill

\section{\uppercase{Introduction}}
\label{sec:introduction}
The dramatic growth in adoption of Neural Networks (NNs) within the last 15 years has sparked the necessity for increased transparency, especially in high-stake applications \citep{eu_ethics_guidelines,agarwal2021responsible}: NNs are indeed considered to be \emph{black boxes} whose rules for outputting predictions are non-interpretable to humans. The quest for transparency considers two different routes: interpretability and explainability \citep{broniatowski2021psychological}. Interpretability is defined as an inherent property of a model, and it is well known that NNs lie very low on the interpretability spectrum, whereas other methods, like decision trees and linear models, are generally considered highly interpretable \citep{james2013introduction}. Explainability, on the other hand, pertains to a series of techniques that can be applied in a \emph{post-hoc} fashion, typically after a model has been trained, in order to gain human-understandable insights on its inner dynamics. The downside of these techniques is that they often approximate complex model dynamics; and issues such as \emph{unfaithfulness} \citep{nauta2023anecdotal,mironicolau2025comprehensive} arise. So, while NNs can often offer accurate solutions and state of the art performance, interpretability has to be achieved through inherently transparent models. When it comes to high-stake applications \cite{rudin2019stop} argues that interpretability is generically a more important property than explainability. Among such high-stake applications, we can find the management of decision making systems that are applied to a large variety of domains, ranging from healthcare to energy management and even financial risk assessment. Interestingly, the recent marriage between Deep and Reinforcement Learning (DRL), has led to the development of several systems that have successfully been applied to these high-stake decision making problems. Examples of such applications go from the aforementioned medical domain \citep{yu2021reinforcement} to particle physics \citep{degrave2022magnetic}, infrastructure management planning \citep{leroy2024imp} and, of course, autonomous driving \citep{sallab2017deep}. Unfortunately, despite working really well in practice, DRL algorithms carry over some of the limitations that characterize NNs, therefore resulting in autonomous agents whose optimal decision making capabilities are not interpretable nor fully explainable. While \cite{glanois2024survey} survey the efforts made by researchers to use interpretable models, such as rules-based methods or formulas, to explain transitions or policies, it is worth mentioning that within the extremely rapidly growing research landscape of Reinforcement Learning (RL), scientific works that are focused on this area are at the present moment rather limited.

\paragraph{Contributions}
In this paper, we aim to make one step forward toward the development of more transparent optimal decision making agents. To achieve this, we rely on a newly introduced learning paradigm that aims to solve typical RL problems from a pure Supervised Learning perspective called Upside Down Reinforcement Learning (UDRL) \citep{schmidhuber2019reinforcement}. While originally designed for being coupled with NNs, we show that an UDRL approach can also be successfully integrated with inherently more interpretable machine learning models, like forests of randomized trees. Despite not being as interpretable as simple decision trees \citep{james2013introduction}, this family of techniques still offers an intrinsic tool for global feature importance, such as mean decrease in impurity, thus offering an edge over NNs in terms of interpretability \citep{ibrahim2019global} and design of transparent autonomous agents.

\section{\uppercase{Upside-Down Reinforcement Learning}}
\label{sec:udrl}

\subsection{The UDRL Framework}
The concept of Upside-Down Reinforcement Learning (UDRL) was first introduced by \citet{schmidhuber2019reinforcement}, and its idea is remarkably straightforward: tackling Reinforcement Learning problems via Supervised Learning (SL) techniques. To understand how this can be achieved, let us introduce the typical Markovian mathematical framework on which popular RL algorithms are based. We define a Markov Decision Process (MDP) \citep{puterman2014markov} as a tuple $\mathcal{M} = \langle \mathcal{S}, \mathcal{A}, \mathcal{P}, \mathcal{R} \rangle$ where $\mathcal{S}$ corresponds to the state space of the environment, $\mathcal{A}$ is the action space modeling all the possible actions available to the agent, $\mathcal{P}: \mathcal{S} \times \mathcal{A} \times \mathcal{S} \rightarrow [0,1] $ is the transition function and $\mathcal{R}: \mathcal{S} \times \mathcal{A} \times \mathcal{S} \rightarrow \mathds{R} $ is the reward function. When interacting with the environment, at each time-step $t$, the RL agent performs action $a_t$ in state $s_t$, and transitions to state $s_{t+1}$ alongside observing reward signal $r_t$. The goal is then to learn a policy $\pi:\mathcal{S}\rightarrow \Delta(\mathcal{A})$ which maximizes the expected discounted sum of rewards $\mathds{E}_\pi[\sum_{k=0}^{\infty}\gamma^{k}r_{t+k}]$, with $\gamma \in [0,1)$. It is well known that when it comes to RL, differently from the Dynamic Programming setting, the transition function $\mathcal{P}$ and the reward function $\mathcal{R}$ are unknown to the agent. To overcome this lack of information, the agent can either learn to predict expected rewards, assuming the model-free RL set-up, or learn to predict $s_{t+1}$ and $r_t$, assuming a model-based RL setting. 

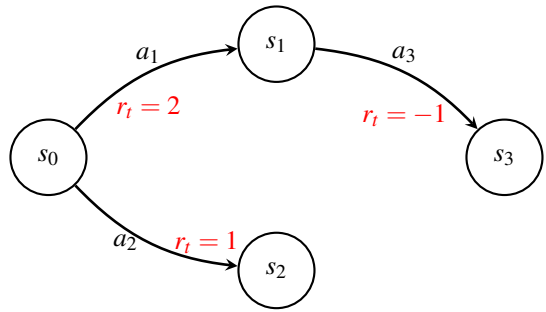
\begin{figure}[h!]
\centering
\begin{tikzpicture}[->, >=stealth, node distance=2.5cm, line width=0.8pt] %
    \node[circle, draw, minimum size=1cm] (s0) at (0,0) {$s_0$};
    \node[circle, draw, minimum size=1cm] (s1) at (3,1.5) {$s_1$};
    \node[circle, draw, minimum size=1cm] (s2) at (3,-1.5) {$s_2$};
    \node[circle, draw, minimum size=1cm] (s3) at (6,0) {$s_3$};

    \draw[->, line width=1pt, bend left=20] (s0) to node[midway, above] {$a_1$} node[midway, below=5pt, red] {$r_t=2$} (s1);
    \draw[->, line width=1pt, bend right=20] (s0) to node[midway, left] {$a_2$} node[midway, right=6pt, red] {$r_t=1$} (s2); 
    \draw[->, line width=1pt, bend left=20] (s1) to node[midway, above] {$a_3$} node[midway, below=8pt, red] {$r_t=-1$} (s3);
\end{tikzpicture}
\caption{A simple MDP whose behavior function $f$ is summarized in Table \ref{tab:behavior_function}.}
\label{fig:mdp}
\end{figure}

In UDRL the agent doesn't do either, and its main goal is learning to predict actions instead. Given a typical RL transition $\tau = \langle s_t, a_t, r_t, s_{t+1}\rangle$, an UDRL agent uses the information contained within $\tau$ to learn to map a certain state $s_t$, and reward $r_t$ to action $a_t$. Formally, this corresponds to learning a function $f(s_t, r_t)=a_t$, a task that can be seen as a classic SL problem where the goal is that of building a function $f:\mathcal{X}\rightarrow\mathcal{Y}$, where $\mathcal{X}$ and $\mathcal{Y}$ are the input and output spaces defining the SL problem one wants to solve. While intuitive and easy to understand, the function defined above does not account for long-horizon rewards, which are crucial for most optimal control problems. Therefore, it needs to be slightly modified. This can be achieved by introducing two additional variables that come with the name of commands: $d_r$ and $d_t$. They correspond to the desired reward $d_r$ the agent wants to achieve within a certain time horizon $d_t$ when being in state $s_t$. Based on this, we can define the ultimate goal of a UDRL agent as that of learning $f(s_t, d_r, d_t)=a_t$. This function $f$ comes with the name of behavior function and, ideally, if learned on a large number of transitions $\tau$, it can be queried in such a way that it can return the action that achieves a certain reward in a certain state in a certain number of time-steps. To better understand the role of this behavior function, let us consider the simple MDP represented in Figure \ref{fig:mdp}, where states are denoted as nodes and edges are represented by the actions $a$ the agent can take, alongside the rewards $r_t$ the agent observes.

A fully trained behavior function that correctly models the dynamics of this MDP is summarized in Table \ref{tab:behavior_function}.

\begin{table}[htbp] 
\caption{A tabular representation of the behavior function $f$ that correctly models the dynamics represented in the MDP visualized in Figure \ref{fig:mdp}. For example, let us consider that the agent is in state $s_0$ and that its goal is that of obtaining a desired reward of $d_r=2$ in one time-step $d_t$, then when queried this function should return $a_1$. Note how the information contained within $s_0$, concatenated to $d_r$ and $d_f$, constitutes a training sample that can be used in an SL setting for predicting $a_1$.}
\centering
\begin{tabular}{>{\columncolor[HTML]{EFEFEF}}lccc}
\toprule
\textbf{State} & \textbf{$d_r$} & \textbf{$d_t$} & \textbf{Action $a$} \\ \midrule
$s_0$ & 2  & 1  & $a_1$ \\
$s_0$ & 1  & 1  & $a_2$ \\
$s_0$ & 1  & 2  & $a_1$ \\
$s_1$ & -1 & 1  & $a_3$ \\
\bottomrule
\end{tabular}
\label{tab:behavior_function}
\end{table}

\paragraph{Related Work}

The first successful application of the UDRL paradigm can be found in the work of \cite{srivastava2019training}, who showed that a UDRL agent controlled by a neural network could outperform popular model-free RL algorithms such as DQN and PPO on both discrete and continuous action space environments. The same paper also highlights the potential of UDRL when it comes to sparse-reward environments. UDRL has been subsequently applied by \cite{ashley2022learning} and \cite{arulkumaran2022all} with the latter work showcasing how this learning paradigm can be successfully extended to a setting of offline RL, imitation learning, and even meta-RL. While the term UDRL has only been introduced very recently, it is worth mentioning that some of its core ideas can be associated with the fields of goal-conditioned or return-conditioned RL \citep{liu2022goal,furuta2021generalized}. In this learning set-up agents need to learn actions constrained to particular goals, which may come in the form of specific rewards in given states, usually from hindsight information. Besides these works, UDRL has, to the best of our knowledge, seldom been applied in other studies if not for comparing it with other similar techniques, such as the Decision Transformer \citep{chen2021decision}. This highlights how UDRL is still an underexplored technique that is still far from expressing its potential.

\section{\uppercase{Methods}}
\label{sec:methods}

In all of the applications above of UDRL, $f$ comes in the form of a neural network, a choice motivated by the recent successes that this family of machine learning techniques has demonstrated when it comes to RL problems modeled by high-dimensional state and action spaces. However, the UDRL formalism introduced by \citet{schmidhuber2019reinforcement} allows for using any machine learning algorithm suitable for solving SL tasks. Yet, to the best of our knowledge, no such study exists. In what follows, we start by exploring whether five different SL algorithms can be effectively coupled with the UDRL framework and therefore provide a valuable alternative to NNs. A brief description of these algorithms, alongside an explanation about why they were chosen, is presented hereafter.

\subsection{Algorithms}

\paragraph{Tree-Based Algorithms} The first two studied algorithms can be categorized under the \emph{tree-based} Reinforcement Learning umbrella and are Random Forests (RFs) \citep{breiman2001random} and Extremely Randomized Trees (ETs) \citep{geurts2006extremely}. Different from neural networks, both methods are nonparametric and build a model in the form of the average prediction of an ensemble of regression trees. Before the advent of neural networks, they both provided a valuable choice of function approximation within the realm of approximate Reinforcement Learning \citep{busoniu2017reinforcement} and have successfully been used for solving several optimal control problems. For example, RFs were successfully coupled with the popular Q-Learning algorithm by \cite{min2022q}, which showed on-par performance to that of Deep-Q Networks on two of the three studied benchmarks. They were also successfully used in a multi-objective setting by \citet{song2024multiobjective}, who optimized multiple objectives in a multistage medical multi-treatment setting. Regarding ETs, the arguably most successful application is that presented by \citet{ernst2005tree}, where ETs were used to construct the popular \textit{Fitted Q Iteration} algorithm for the batch RL setting. ETs have since then been successfully used for solving several RL tasks, ranging from the design of clinical trials \citep{ernst2006clinical, zhao2009reinforcement} to the management of intensive care units \citep{prasad2017reinforcement} and even autonomous driving \citep{mirchevska2017reinforcement}. Among the benefits of this family of techniques \cite{wehenkel2006ensembles} highlight their universal approximation consistency, alongside their interpretability, two properties which, in the UDRL context, could allow learning an interpretable and transparent behavior function $f$.

\paragraph{Boosting Algorithms}
The third and fourth studied behavior functions are ensemble methods that fall within the \emph{boosting} family of algorithms. Differently from the aforementioned tree-based algorithms, which construct the trees independently and in parallel, boosting techniques build the trees sequentially, with the main goal of having each new tree learn from the errors of the previous one. While several boosting algorithms exist, in this paper we consider the two arguably most popular ones: AdaBoost \citep{freund1996experiments} and XGBoost \citep{friedman2001greedy}. Both of these algorithms have successfully been used in RL \citep{brukhim2022boosting}, although their RL applications are scarcer compared to RFs and ETs. Yet, we believe that within the UDRL setting, they both could be used for effectively learning the behavior function $f$. As was described in Table \ref{tab:behavior_function}, a typical UDRL behavior function is trained on data that comes in tabular form, a data type on which boosting algorithms typically perform very well and can even outperform NNs \citep{shwartz2022tabular}.

\paragraph{K-Nearest Neighbour}
The final considered algorithm is K-Nearest Neighbour (KNN) \citep{cover1967nearest}, a popular clustering method rooted in instance-based learning. While its applications when it comes to RL are rather limited, made some exceptions for its integration within some model-free algorithms \citep{martin2009k, shah2018q}, this algorithm was chosen as it provides a straightforward non-parametric baseline that can be effectively used for assessing the performance of the previously introduced tree-based and boosting methods.

\subsection{Environments}
\label{subsec:environments}

We evaluate the performance of the abovementioned algorithms on three popular optimal control benchmarks of increasing difficulty. The benchmarks are \texttt{CartPole} \citep{barto1983neuronlike}, \texttt{Acrobot} \citep{sutton1995generalization} and \texttt{Lunar Lander} and are all provided by the OpenAI-Gym package. Each environment is summarized in Table \ref{tab:MDPEnvironments}, where we describe its key components, by specifically focusing on the state space $\mathcal{S}$ representation, as this information will play an important role in Section \ref{sec:interpetability} of the paper. 

\begin{table*}[h!]
\centering
\scriptsize
\caption{An overall description of the \texttt{CartPole}, \texttt{Acrobot}, and \texttt{Lunar Lander} environments used in our experiments. For each environment, we report information regarding its state $\mathcal{S}$ and $\mathcal{A}$ action spaces, together with the rewards that the agent receives while interacting.}
\begin{tabular}{>{\columncolor[HTML]{EFEFEF}}l p{4cm} p{4cm} p{4cm}} 
\toprule
\textbf{Component} & \texttt{CartPole} & \texttt{Acrobot} & \texttt{Lunar Lander} \\ \midrule
\textbf{States} \(\mathcal{S}\) & 
\begin{tabular}[t]{@{}l@{}} 
Continuous 4D vector: \\ 
- \(x\): Cart Position  \\ 
- \(\dot{x}\): Cart Velocity  \\ 
- \(\theta\): Pole Angle \\ 
- \(\dot{\theta}\): Pole Angular Velocity
\end{tabular} & 
\begin{tabular}[t]{@{}l@{}} 
Continuous 6D vector: \\ 
- \(\sin{\theta_1}\): Sine of the first link \\ 
- \(\cos{\theta_1}\): Cosine of the first link \\ 
- \(\sin{\theta_2}\): Sine of the second link \\ 
- \(\cos{\theta_2}\): Cosine of the second link \\ 
- \(\dot{\theta}_1\): Angular velocity of first link \\ 
- \(\dot{\theta}_2\): Angular velocity of second link
\end{tabular} & 
\begin{tabular}[t]{@{}l@{}} 
Continuous 8D vector: \\ 
- \(x\): X position \\ 
- \(y\): Y position \\ 
- \(\dot{x}\): X linear velocity \\ 
- \(\dot{y}\): Y linear velocity \\ 
- \(\theta\): Angle \\ 
- \(\dot{\theta}\): Angular velocity \\ 
- Left and right leg contact points
\end{tabular} \\ \midrule

\textbf{Actions} \(\mathcal{A}\) & 
\begin{tabular}[t]{@{}l@{}} 
Discrete: \\ 
0: Push left \\ 
1: Push right 
\end{tabular} & 
\begin{tabular}[t]{@{}l@{}} 
Discrete: \\ 
0: Torque left \\ 
1: No torque \\ 
2: Torque right 
\end{tabular} & 
\begin{tabular}[t]{@{}l@{}} 
Discrete: \\ 
0: Do nothing \\ 
1: Fire left engine \\ 
2: Fire main engine \\ 
3: Fire right engine 
\end{tabular} \\ \midrule

\textbf{Rewards} $r_t$ & 
+1 for every step taken & 
-1 for every step until the goal is reached & 
\begin{tabular}[t]{@{}l@{}} 
+100 for successful landing \\ 
-100 for crash landing \\ 
Reward for firing engines and leg contact
\end{tabular} \\ \bottomrule
\end{tabular}
\label{tab:MDPEnvironments}
\end{table*}

\subsection{Experimental Setup}
\label{sec:experimental_setup}

We train each behavior function on the aforementioned optimal control tasks five times with five different random seeds by following the algorithmic guidelines described by \cite{srivastava2019training}, the full pseudo-code of our UDRL implementation can be found in the Appendix \ref{algo:udrl}. Each algorithm is allocated a budget of 500 training episodes, as this is a number of episodes that are known to be sufficient for resulting in successful learning for most popular DRL algorithms. When it comes to the NN, the behavior function comes in the form of a multi-layer perceptron with three hidden layers, activated by ReLU activation functions. The size of each hidden layer is $64$ nodes and the popular Adam algorithm with a learning rate of $0.001$ is used for optimization. The network is implemented in the \texttt{PyTorch} deep-learning framework \citep{paszke2017automatic}. All other behavior functions are implemented with the \texttt{Scikit-Learn} package \citep{pedregosa2011scikit} and we adopt the default hyperparameters that are provided by the library. Exploration of all agents is governed via the popular $\epsilon$-greedy strategy with the value of $\epsilon$ set constant to $0.2$ throughout training. Lastly, as is common practice, all behavior functions are trained on UDRL transitions that are stored in an Experience-Replay memory buffer which has a capacity of 700.

\section{\uppercase{Results}}
\label{sec:results}

We now provide an analysis of the performance of the six different tested behavior functions $f$. Our analysis is twofold: first, we focus on the performance obtained at training time, while later we focus on the performance obtained at inference time. 

\begin{figure*}[htbp]
    \centering
    \begin{subfigure}{0.32\textwidth}
        \centering
        \begin{tikzpicture}
            \begin{axis}[
                title={\texttt{CartPole}},
                width=\linewidth,
                height=\linewidth,
                xlabel={Episode},
                ylabel={Reward},
                ymin=0, ymax=200,
                xtick={0, 250, 500},
                ytick={0, 100, 200},
                grid=both,
                major grid style={line width=0.8pt, draw=gray!70},
                minor grid style={line width=0.4pt, draw=gray!50},
                ]
                \addplot+[smooth, color=blue, mark=o, very thick, mark repeat=50,mark phase=50] table [col sep=comma, x=episode, y=NN_mean_smoothed] {Results/CartPole-v0.csv};
                \addplot+[smooth, color=green, mark=star, thick, mark repeat=50,mark phase=50] table [col sep=comma, x=episode, y=RF_mean_smoothed] {Results/CartPole-v0.csv};
                \addplot+[smooth, color=red, mark=x, very thick, mark repeat=50,mark phase=50] table [col sep=comma, x=episode, y=ET_mean_smoothed] {Results/CartPole-v0.csv};
                \addplot+[smooth, color=orange, mark=asterisk, very thick, mark repeat=50,mark phase=50] table [col sep=comma, x=episode, y=AdaBoost_mean_smoothed] {Results/CartPole-v0.csv};
                \addplot+[smooth, color=cyan, mark=triangle, very thick, mark repeat=50,mark phase=50] table [col sep=comma, x=episode, y=XGBoost_mean_smoothed] {Results/CartPole-v0.csv};
                \addplot+[smooth, color=purple, mark=diamond, very thick, mark repeat=50,mark phase=150] table [col sep=comma, x=episode, y=KNN_mean_smoothed] {Results/CartPole-v0.csv};
            \end{axis}
        \end{tikzpicture}
    \end{subfigure}%
    \begin{subfigure}{0.32\textwidth}
        \centering
        \begin{tikzpicture}
            \begin{axis}[
                title={\texttt{Acrobot}},
                width=\linewidth,
                height=\linewidth,
                xlabel={Episode},
                ymin=-500, ymax=-50,
                xtick={0, 250, 500},
                ytick={-500, -250, -50},
                grid=both,
                major grid style={line width=0.8pt, draw=gray!70},
                minor grid style={line width=0.4pt, draw=gray!50},
                ]
                \addplot+[smooth, color=blue, mark=o, very thick, mark repeat=50,mark phase=50] table [col sep=comma, x=episode, y=NN_mean_smoothed] {Results/Acrobot-v1.csv};
                \addplot+[smooth, color=green, mark=star, thick, mark repeat=50,mark phase=50] table [col sep=comma, x=episode, y=RF_mean_smoothed] {Results/Acrobot-v1.csv};
                \addplot+[smooth, color=red, mark=x, very thick, mark repeat=50,mark phase=50] table [col sep=comma, x=episode, y=ET_mean_smoothed] {Results/Acrobot-v1.csv};
                \addplot+[smooth, color=orange, mark=asterisk, very thick, mark repeat=50,mark phase=50] table [col sep=comma, x=episode, y=AdaBoost_mean_smoothed] {Results/Acrobot-v1.csv};
                \addplot+[smooth, color=cyan, mark=triangle, very thick, mark repeat=50,mark phase=50] table [col sep=comma, x=episode, y=XGBoost_mean_smoothed] {Results/Acrobot-v1.csv};
                \addplot+[smooth, color=purple, mark=diamond, very thick, mark repeat=50,mark phase=150] table [col sep=comma, x=episode, y=KNN_mean_smoothed] {Results/Acrobot-v1.csv};
            \end{axis}
        \end{tikzpicture}
    \end{subfigure}%
    \begin{subfigure}{0.32\textwidth}
        \centering
        \begin{tikzpicture}
            \begin{axis}[
                title={\texttt{Lunar Lander}},
                width=\linewidth,
                height=\linewidth,
                xlabel={Episode},
                ymin=-350, ymax=0,
                xtick={0, 250, 500},
                ytick={-250, -150, -50},
                grid=both,
                major grid style={line width=0.8pt, draw=gray!70},
                minor grid style={line width=0.4pt, draw=gray!50},
                ]
                \addplot+[smooth, color=blue, mark=o, very thick, mark repeat=50,mark phase=50] table [col sep=comma, x=episode, y=NN_mean_smoothed] {Results/LunarLander-v2.csv};
                \addplot+[smooth, color=green, mark=star, thick, mark repeat=50,mark phase=50] table [col sep=comma, x=episode, y=RF_mean_smoothed] {Results/LunarLander-v2.csv};
                \addplot+[smooth, color=red, mark=x, very thick, mark repeat=50,mark phase=50] table [col sep=comma, x=episode, y=ET_mean_smoothed] {Results/LunarLander-v2.csv};
                \addplot+[smooth, color=orange, mark=asterisk, very thick, mark repeat=50,mark phase=50] table [col sep=comma, x=episode, y=AdaBoost_mean_smoothed] {Results/LunarLander-v2.csv};
                \addplot+[smooth, color=cyan, mark=triangle, very thick, mark repeat=50,mark phase=50] table [col sep=comma, x=episode, y=XGBoost_mean_smoothed] {Results/LunarLander-v2.csv};
                \addplot+[smooth, color=purple, mark=diamond, very thick, mark repeat=50,mark phase=150] table [col sep=comma, x=episode, y=KNN_mean_smoothed] {Results/LunarLander-v2.csv};
            \end{axis}
        \end{tikzpicture}
    \end{subfigure}
    
    \vspace{0.5cm} 
    
    \begin{minipage}{\textwidth}
        \centering
        \begin{tikzpicture}
            \begin{axis}[
                axis y line=none,
                axis x line=none,
                width=0.9\textwidth,
                height=0.2\textwidth,
                legend columns=3, 
                legend style={at={(0.5,1)}, anchor=north, column sep=1ex, font=\footnotesize, legend cell align=left},
                ]
                \addplot+[color=blue, mark=o, thick] coordinates {(0,0)}; \addlegendentry{NN}
                \addplot+[color=green, mark=star, thick] coordinates {(0,0)}; \addlegendentry{RF}
                \addplot+[color=red, mark=x, thick] coordinates {(0,0)}; \addlegendentry{ET}
                \addplot+[color=orange, mark=asterisk, thick] coordinates {(0,0)}; \addlegendentry{AdaBoost}
                \addplot+[color=cyan, mark=triangle, thick] coordinates {(0,0)}; \addlegendentry{XGBoost}
                \addplot+[color=purple, mark=diamond, thick] coordinates {(0,0)}; \addlegendentry{KNN}
            \end{axis}
        \end{tikzpicture}
    \end{minipage}

    \caption{Comparison of the performance of the six different tested behavior functions (NN, RF, ET, KNN, AdaBoost, and XGBoost) on the three OpenAI Gym environments: \texttt{CartPole}, \texttt{Acrobot}, and \texttt{Lunar Lander}. The results are shown in terms of rewards per episode and are averaged over five different training runs.}
    \label{fig:comparison}
\end{figure*}
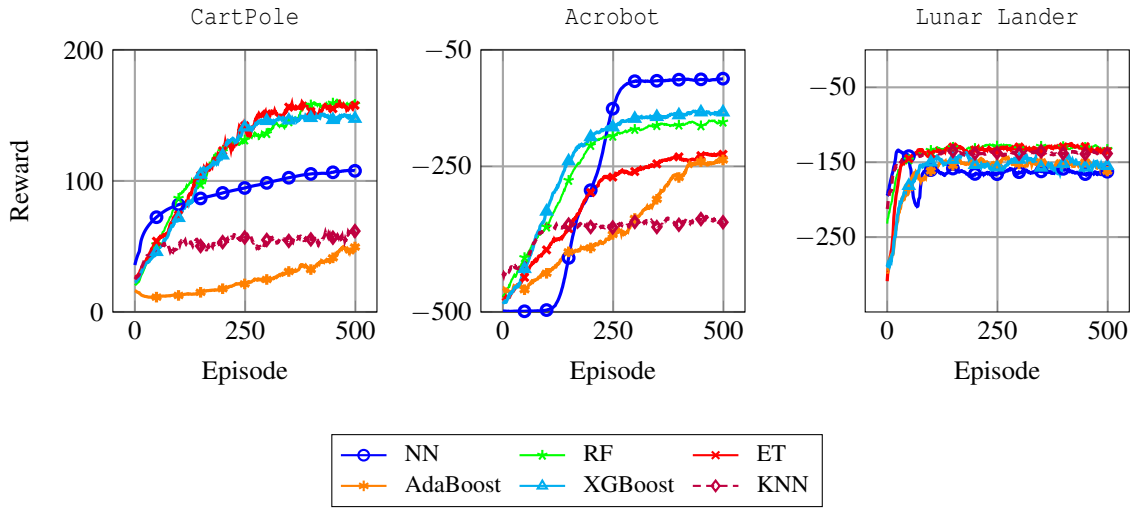

\paragraph{Training} Our training curves are reported in Figure \ref{fig:comparison} where for each algorithm we show the average reward obtained per training episode over five different training runs. We can start by observing that on the \texttt{CartPole} task the best-performing behavior functions appear to be RFs, ETs, and XGBoost, as they all obtain training rewards of $\approx 150$. The performance of these ensemble methods is therefore better than that obtained by the NN, which, despite showing some faster initial training, obtains a final reward of $\approx 100$. The KNN and AdaBoost behavior functions on the other hand show less promising results as their performance only marginally improves over training. When it comes to the \texttt{Acrobot} task we can observe that this time the best performing algorithm is the NN, as it converges to a training reward of $\approx -150$. Among the tree-based methods, the RF behavior function is the one performing best, obtaining training rewards of $\approx -200$. A similar performance is also obtained by the XGBoost algorithm. ETs and AdaBoost eventually perform very similarly although the ET al algorithm converges significantly faster. Similar to what was observed on the \texttt{CartPole} task, the worst performing algorithm is again the KNN behavior function. On the \texttt{Lunar Lander} environment we can observe that all tested behavior functions are able to improve their decision making across episodes. Therefore, suggesting that there is no algorithm that significantly outperforms all of the others.

\paragraph{Inference} While the results presented above are promising and possibly suggest that in the UDRL context the behavior function $f$ does not have to necessarily come in the form of a neural network, it is essential to investigate how general and robust this behavior function is. Note that, as was described in Section \ref{sec:experimental_setup}, the results presented in Figure \ref{fig:comparison} are computed at training time, which means that the reported performance is representative of the $\epsilon$-greedy exploration strategy that the algorithms follow whilst training. As a result, the performance presented in the above-mentioned training curves is only partially representative of the final performance of the different algorithms. To quantitatively assess how well the trained behavior functions have actually mastered the three different optimal control environments, one needs to query these functions. Recall, that based on the formalism explained in Section \ref{sec:udrl}, the ultimate goal of UDRL is that of learning a behavior function $f$, that once queried with a state coming from the environment $s_t$ can return action $a_t$ achieving desired reward $d_r$ in time horizon $d_t$. Therefore, to test the inference capabilities of the different behavior functions it is crucial to define $d_r$ and $d_t$. For the \texttt{CartPole} task this is straightforward as it is well known that given the episodic nature of the problem, the maximum cumulative reward an agent can obtain is that of $200$. It follows that all behavior functions are queried with $d_r$ and $d_t$ set to $200$. When it comes to the \texttt{Acrobot} and \texttt{Lunar Lander} environments, which differently from \texttt{CartPole} are non-episodic tasks, setting $d_r$ and $d_t$ requires some extra care. As one wants to avoid queering the behavior function with desired commands $c$ that are unrealistic and not supported by the environment, we decided to set $d_r$ and $d_t$ to the most common used values that were used throughout the last $100$ episodes, as we believe they are fairly representative of the final performance of the algorithms. For each behavior function and environment, we report the exact form of $f(s_t, d_r, d_t)$ in Table \ref{tab:inference_parameters}.

\begin{table}[htbp] 
\caption{The form of the behavior function $f$ that is used at inference time by each algorithm on the three tested environments.}
\centering
\tiny
\begin{tabular}{>{\columncolor[HTML]{EFEFEF}}lcccc}
\toprule
\textbf{Behaviour Function} & \texttt{CartPole}   & \texttt{Acrobot}   & \texttt{Lunar Lander}   \\ \midrule
Neural Network     & $f(\cdot, 200, 200)$ & $f(\cdot, -63, 64)$ & $f(\cdot, 49, 101)$ &      \\ 
Random Forest      & $f(\cdot, 200, 200)$& $f(\cdot, -79, 82)$& $f(\cdot, 57, 102)$ &      \\  
Extra-Trees     & $f(\cdot, 200, 200)$& $f(\cdot, -75, 77)$&  $f(\cdot, 34, 118)$&      \\     
AdaBoost     &$f(\cdot, 200, 200)$ & $f(\cdot, -77, 81)$&  $f(\cdot, 45, 118)$     \\ 
XGBoost     & $f(\cdot, 200, 200)$& $f(\cdot, -114, 120)$ & $f(\cdot, 136, 389)$ &      \\ 
KNN    & $f(\cdot, 200, 200)$& $f(\cdot, -129, 132)$ &  $f(\cdot, 62, 112)$&      \\ 
\bottomrule
\end{tabular}
\label{tab:inference_parameters}
\end{table}

The results obtained at inference time are presented in Table \ref{tab:inference_results} where we report the average cumulative reward that is obtained by the behavior function presented in Table \ref{tab:inference_parameters} which is queried 100 times. The results of the best performing behavior function are presented in a green cell, whereas the ones of the second best performing one are reported in a yellow cell. On the \texttt{CartPole} task we can observe that the overall best performing behavior function is the one coming in the form of a NN, followed by the one modeled by the XGBoost algorithm, as both behavior functions fully solve the presented task by obtaining a final cumulative reward of $\approx 200$. The performance of the tree-based algorithms is slightly lower, and a bit more unstable as shown by the larger standard deviations, however, it can still be stated that both algorithms have learned how to balance the cart successfully. Good performance was also obtained by the AdaBoost algorithm, albeit not on par with that of the aforementioned tree-based algorithms. Finally, the KNN behavior function remains the worst performing algorithm as was already highlighted when discussing the first plot presented in Figure \ref{fig:comparison}, where the algorithm barely improved its decision making over time. When it comes to the \texttt{Acrobot} task, we can again see that the best performing behavior function is that modeled by a NN which obtains final testing rewards of $\approx -75$. RFs, ETs, and XGBoost all perform similarly, although their final performance is slightly worse compared to that of NNs given final testing rewards of $\approx -100$. Again, we can note that the results obtained by the AdaBoost algorithm are not on-par with those obtained by the tree-based methods and that the KNN algorithm is overall the worse performing algorithm. On the arguably most complicated \texttt{Lunar Lander} environment, we can see that the best performing behavior function is the one modeled via RFs, which obtains a final performance of $\approx -54$. Similarly to what was observed on the \texttt{CartPole} task the second best performing algorithm is XGBoost, and perhaps surprisingly the third best performing algorithm is AdaBoost. Arguably even more surprising is the fact that on this task the KNN behavior function managed to actually improve its performance over time, and that the behavior function modeled by the NN is the overall worst performing algorithm.

\begin{table*}[htbp]
\caption{The results of the six tested behavior functions obtained at inference time, namely when the behavior function $f$ gets queried with the desired reward $d_t$ and desired time horizon $d_t$ commands described in Table \ref{tab:inference_parameters}. The best performing algorithm is reported in a green cell, whereas the second best performing algorithm is reported in a yellow cell.}
\centering
\begin{tabular}{>{\columncolor[HTML]{EFEFEF}}lcccc}
\textbf{Behavior Function} & \texttt{CartPole}   & \texttt{Acrobot}   & \texttt{Lunar Lander}   \\ \midrule
Neural Network     & \cellcolor{green!25}$199.93 \pm 0.255$ &\cellcolor{green!25}$-75.00 \pm 15.36$ & $-157.04 \pm 71.26$ &      \\ 
Random Forest      &$188.25 \pm 13.82$ &\cellcolor{yellow!25}$-100.05 \pm 62.80$ &\cellcolor{green!25}$-54.74 \pm 96.22$ &      \\ 
Extra-Trees     &$181.56 \pm 31.63$ & $ -100.00 \pm 93.72$& $-127.79 \pm49.19$&      \\
AdaBoost     & $168.61\pm31.02$ & $-109.00\pm63.70$& $-108.26\pm80.71$ &      \\
XGBoost     &\cellcolor{yellow!25}$199.27\pm4.06$&$-100.00\pm34.71$ &\cellcolor{yellow!25}$-76.96\pm89.69$ &      \\
KNN    & $85.11\pm50.30$ &$-160.50 \pm 36.66$  & $-128.47\pm47.49 $&      \\
\bottomrule
\end{tabular}
\label{tab:inference_results}
\end{table*}

\section{\uppercase{Interpretability Gains}}
\label{sec:interpetability}

While RFs and ETs did not significantly outperform their NN counterparts in terms of final performance, it is worth noting that this family of techniques has much more to offer regarding interpretability than NNs as it allows us to get global explanations for the behavior function, which in the case of NNs is not straightforward. In this section, we aim to shed some light on what makes the aforementioned tree-based behavior functions able to solve the optimal control tasks described in Section \ref{subsec:environments} almost optimally. To this end, we rely on the work of \cite{louppe2013understanding}: for different states of the tested environments, we compute feature importance scores that are estimated as mean impurity decrease. We compute these scores at inference time, based on the experimental protocol that was discussed in the second paragraph of Section \ref{sec:results}. We present a qualitative analysis that shows how for different states of the environments, sampled at different stages of the RL episode, the feature importance scores change and can highlight which features of the state space $\mathcal{S}$ are the most relevant for the behavior function. 

\begin{figure*}[htbp]
    \centering
    \begin{subfigure}{0.3\textwidth}
        \centering
        \begin{subfigure}{1\textwidth}
            \includegraphics[width=\textwidth]{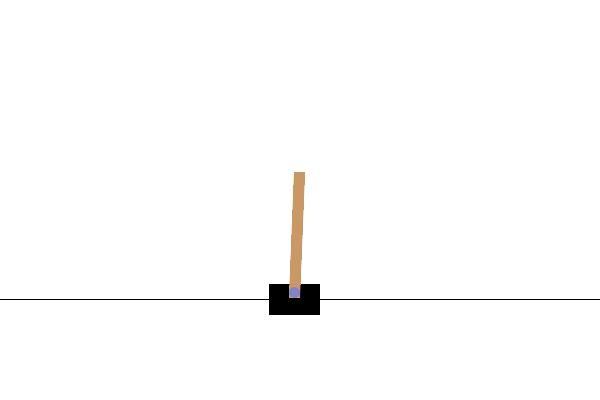}
        \end{subfigure} 
        \begin{subfigure}{1\textwidth}
            \centering
            \begin{tikzpicture}
                \begin{axis}[
                    ylabel={Importance},
                    grid=both,
                    width=\textwidth,
                    height=0.9\textwidth,
                    xtick=data,
                    xticklabels={\(x\), \(\dot{x}\),\(\theta\),\(\dot{\theta}\), $d_r$, $d_t$},
                    x tick label style={rotate=45, anchor=east},
                    enlarge x limits=0.05,
                    ybar,
                    bar width=10pt,
                    error bars/y dir=both,
                    error bars/y explicit
                ]
                \addplot[fill=red, draw=red] table {Plots/feature_viz/CartPole-v0/RandomForestClassifier/info_0.dat};
                \end{axis}
            \end{tikzpicture}
        \end{subfigure}
    \end{subfigure}
    \begin{subfigure}{0.3\textwidth}
        \centering
        \begin{subfigure}{1\textwidth}
            \includegraphics[width=\textwidth]{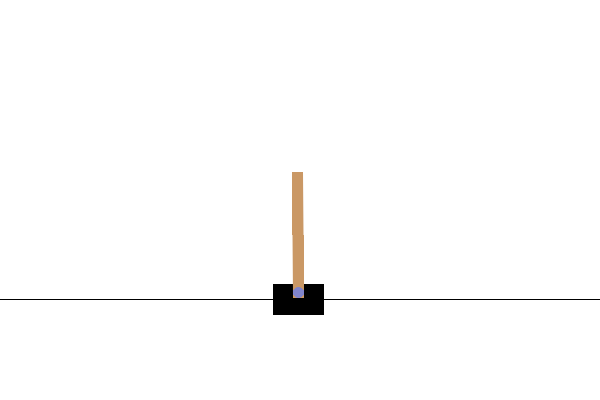}
        \end{subfigure} 
        \begin{subfigure}{1\textwidth}
            \centering
            \begin{tikzpicture}
                \begin{axis}[
                    grid=both,
                    width=\textwidth,
                    height=0.9\textwidth,
                    xtick=data,
                    xticklabels={\(x\), \(\dot{x}\),\(\theta\),\(\dot{\theta}\), $d_r$, $d_t$},
                    x tick label style={rotate=45, anchor=east},
                    enlarge x limits=0.05,
                    ybar,
                    bar width=10pt,
                    error bars/y dir=both,
                    error bars/y explicit
                ]
                \addplot[fill=red, draw=red] table {Plots/feature_viz/CartPole-v0/RandomForestClassifier/info_1.dat};
                \end{axis}
            \end{tikzpicture}
        \end{subfigure}
    \end{subfigure}
    \begin{subfigure}{0.3\textwidth}
        \centering
        \begin{subfigure}{1\textwidth}
            \includegraphics[width=\textwidth]{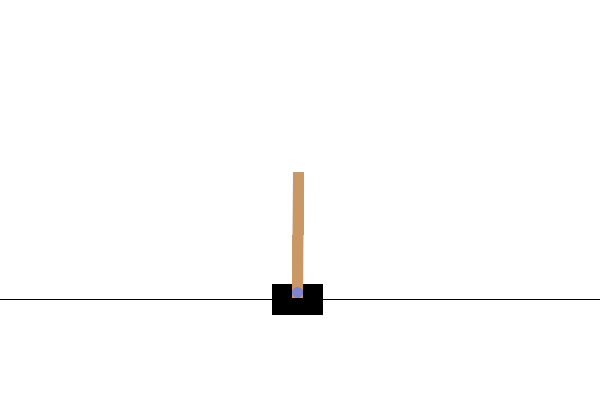}
        \end{subfigure} 
        \begin{subfigure}{1\textwidth}
            \centering
            \begin{tikzpicture}
                \begin{axis}[
                    grid=both,
                    width=\textwidth,
                    height=0.9\textwidth,
                    xtick=data,
                    xticklabels={\(x\), \(\dot{x}\),\(\theta\),\(\dot{\theta}\), $d_r$, $d_t$},
                    x tick label style={rotate=45, anchor=east},
                    enlarge x limits=0.05,
                    ybar,
                    bar width=10pt,
                    error bars/y dir=both,
                    error bars/y explicit
                ]
                \addplot[fill=red, draw=red] table {Plots/feature_viz/CartPole-v0/RandomForestClassifier/info_2.dat};
                \end{axis}
            \end{tikzpicture}
        \end{subfigure}
    \end{subfigure}
    \caption{Feature importance scores coming from a trained RF behavior function computed for three different states of the \texttt{CartPole} environment.}
        \label{fig:cartpole_fi}
\end{figure*}
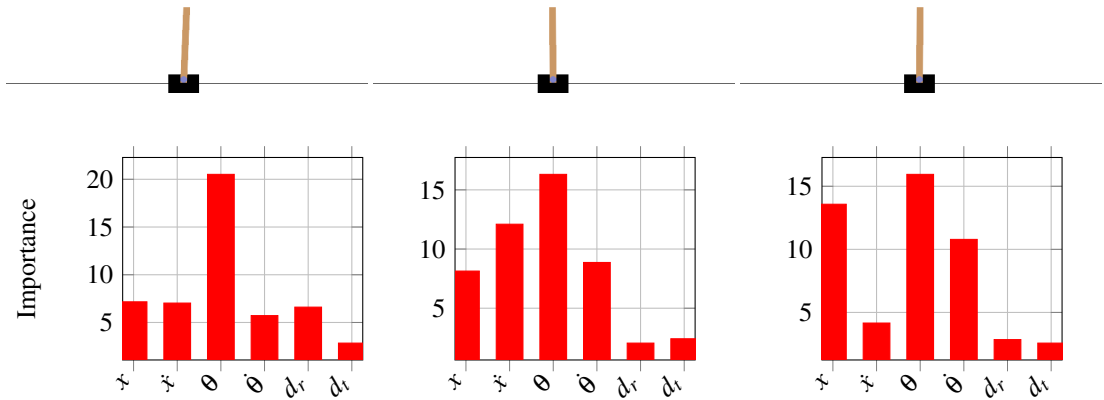

We start by discussing the scores obtained on the \texttt{CartPole} task by an RF behavior function. These results are summarized in Figure \ref{fig:cartpole_fi}. Overall we can observe that out of the four different features modeling the state space, the highest importance score throughout an episode is always associated with the pole angular velocity $\theta$ feature, as demonstrated by all three histogram plots of Figure \ref{fig:cartpole_fi}. This feature appears to play an even more significant role when the pole on top of the cart is in an out-of-balance position (see first plot of Figure \ref{fig:cartpole_fi}). We can also observe that at the very beginning of the episode (last plot of Figure \ref{fig:cartpole_fi}), high-importance scores are associated with the feature denoting the position of the cart $x$, although the more the agent interacts with the environment, the less important this feature becomes.

\begin{figure*}[htb!]
    \centering
    \begin{subfigure}{0.3\textwidth}
        \centering
        \begin{subfigure}{1\textwidth}
            \includegraphics[width=\textwidth]{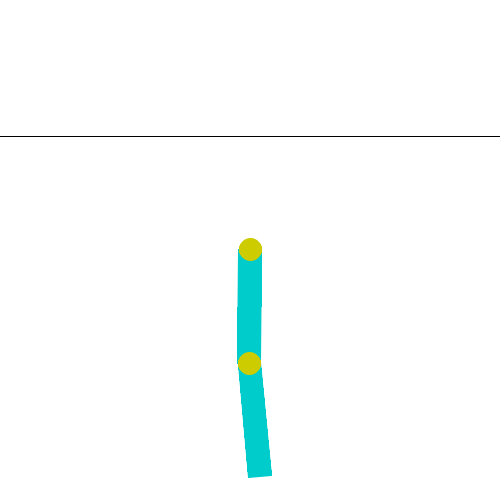}
        \end{subfigure} 
        \begin{subfigure}{1\textwidth}
            \centering
            \begin{tikzpicture}
                \begin{axis}[
                    ylabel={Importance},
                    grid=both,
                    width=\textwidth,
                    height=0.9\textwidth,
                    xtick=data,
                    xticklabels={\(\sin{\theta_1}\),\(\cos{\theta_1}\),\(\sin{\theta_2}\),\(\cos{\theta_2}\),\(\dot{\theta}_1\),\(\dot{\theta}_2\),$d_r$, $d_t$},
                    x tick label style={rotate=45, anchor=east},
                    enlarge x limits=0.05,
                    ybar,
                    bar width=10pt,
                    error bars/y dir=both,
                    error bars/y explicit
                ]
                \addplot[fill=blue, draw=blue] table {Plots/feature_viz/Acrobot-v1/RandomForestClassifier/info_0.dat};
                \end{axis}
            \end{tikzpicture}
        \end{subfigure}
    \end{subfigure}
    \begin{subfigure}{0.3\textwidth}
        \centering
        \begin{subfigure}{1\textwidth}
            \includegraphics[width=\textwidth]{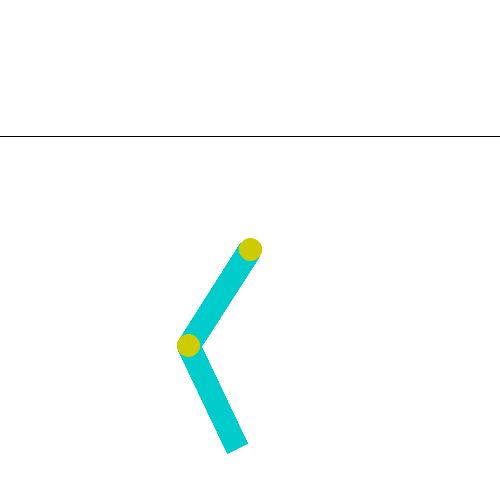}
        \end{subfigure} 
        \begin{subfigure}{1\textwidth}
            \centering
            \begin{tikzpicture}
                \begin{axis}[
                    grid=both,
                    width=\textwidth,
                    height=0.9\textwidth,
                    xtick=data,
                    xticklabels={\(\sin{\theta_1}\),\(\cos{\theta_1}\),\(\sin{\theta_2}\),\(\cos{\theta_2}\),\(\dot{\theta}_1\),\(\dot{\theta}_2\),$d_r$, $d_t$},
                    x tick label style={rotate=45, anchor=east},
                    enlarge x limits=0.05,
                    ybar,
                    bar width=10pt,
                    error bars/y dir=both,
                    error bars/y explicit
                ]
                \addplot[fill=blue, draw=blue] table {Plots/feature_viz/Acrobot-v1/RandomForestClassifier/info_1.dat};
                \end{axis}
            \end{tikzpicture}
        \end{subfigure}
    \end{subfigure}
    \begin{subfigure}{0.3\textwidth}
        \centering
        \begin{subfigure}{1\textwidth}
            \includegraphics[width=\textwidth]{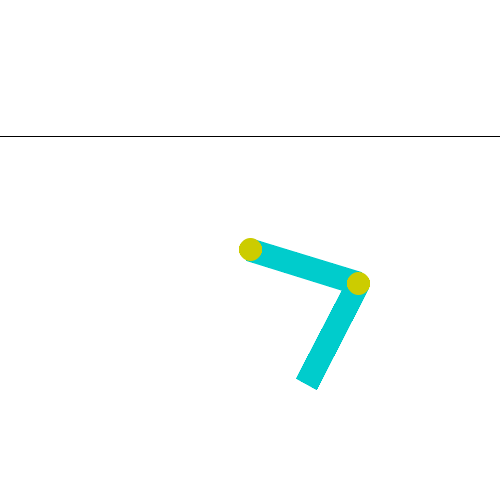}
        \end{subfigure} 
        \begin{subfigure}{1\textwidth}
            \centering
            \begin{tikzpicture}
                \begin{axis}[
                    grid=both,
                    width=\textwidth,
                    height=0.9\textwidth,
                    xtick=data,
                    xticklabels={\(\sin{\theta_1}\),\(\cos{\theta_1}\),\(\sin{\theta_2}\),\(\cos{\theta_2}\),\(\dot{\theta}_1\),\(\dot{\theta}_2\),$d_r$, $d_t$},
                    x tick label style={rotate=45, anchor=east},
                    enlarge x limits=0.05,
                    ybar,
                    bar width=10pt,
                    error bars/y dir=both,
                    error bars/y explicit
                ]
                \addplot[fill=blue, draw=blue] table {Plots/feature_viz/Acrobot-v1/RandomForestClassifier/info_2.dat};
                \end{axis}
            \end{tikzpicture}
        \end{subfigure}
    \end{subfigure}
    \caption{Feature importance scores coming from a trained ET behavior function computed for three different states of the \texttt{Acrobot} environment.}
    \label{fig:acrobot_fi}
\end{figure*}

On the \texttt{Acrobot} environment, where we report feature importance scores for the ET behavior function, we can observe from the results presented in Figure \ref{fig:acrobot_fi} that the most important features are by far the ones representing the two angularComparison with State of the Art velocities of the two links ($\dot{\theta}_1$ and $\dot{\theta}_2$). These two features are consistently associated to the two highest importance scores, no matter at which stage of the episode the agent is.

\begin{figure*}[htbp]
    \centering
    \begin{subfigure}{0.3\textwidth}
        \centering
        \begin{subfigure}{1\textwidth}
            \includegraphics[width=\textwidth]{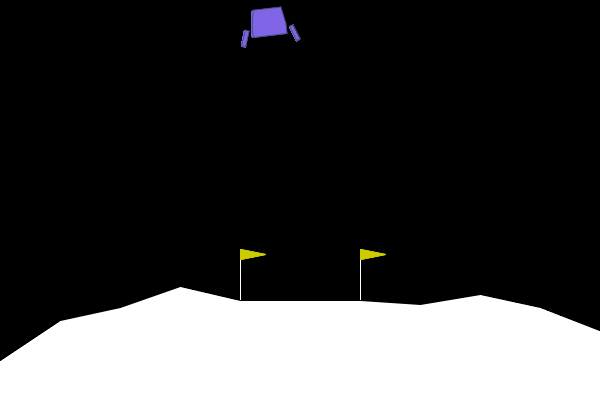}
        \end{subfigure} 
        \begin{subfigure}{1\textwidth}
            \centering
            \begin{tikzpicture}
                \begin{axis}[
                    grid=both,
                    width=\textwidth,
                    height=0.9\textwidth,
                    xtick=data,
                    xticklabels={
                        \(x\),\(y\),\(\dot{x}\),\(\dot{y}\),\(\theta\),\(\dot{\theta}\),$lc$,$rc$,$d_r$, $d_t$},
                    x tick label style={rotate=45, anchor=east},
                    enlarge x limits=0.05,
                    ybar,
                    bar width=10pt,
                    error bars/y dir=both,
                    error bars/y explicit
                ]
                \addplot[fill=red, draw=red] table {Plots/feature_viz/LunarLander-v2/RandomForestClassifier/info_0.dat};
                \end{axis}
            \end{tikzpicture}
        \end{subfigure}
    \end{subfigure}
    \begin{subfigure}{0.3\textwidth}
        \centering
        \begin{subfigure}{1\textwidth}
            \includegraphics[width=\textwidth]{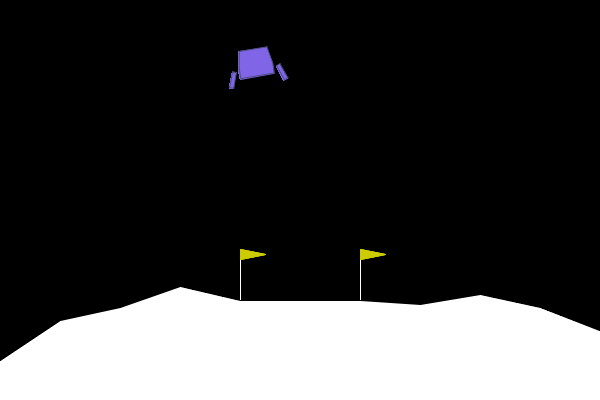}
        \end{subfigure} 
        \begin{subfigure}{1\textwidth}
            \centering
            \begin{tikzpicture}
                \begin{axis}[
                    ylabel={Importance},
                    grid=both,
                    width=\textwidth,
                    height=0.9\textwidth,
                    xtick=data,
                    xticklabels={
                        \(x\),\(y\),\(\dot{x}\),\(\dot{y}\),\(\theta\),\(\dot{\theta}\),$lc$,$rc$,$d_r$, $d_t$},
                    x tick label style={rotate=45, anchor=east},
                    enlarge x limits=0.05,
                    ybar,
                    bar width=10pt,
                    error bars/y dir=both,
                    error bars/y explicit
                ]
                \addplot[fill=red, draw=red] table {Plots/feature_viz/LunarLander-v2/RandomForestClassifier/info_1.dat};
                \end{axis}
            \end{tikzpicture}
        \end{subfigure}
    \end{subfigure}
    \begin{subfigure}{0.3\textwidth}
        \centering
        \begin{subfigure}{1\textwidth}
            \includegraphics[width=\textwidth]{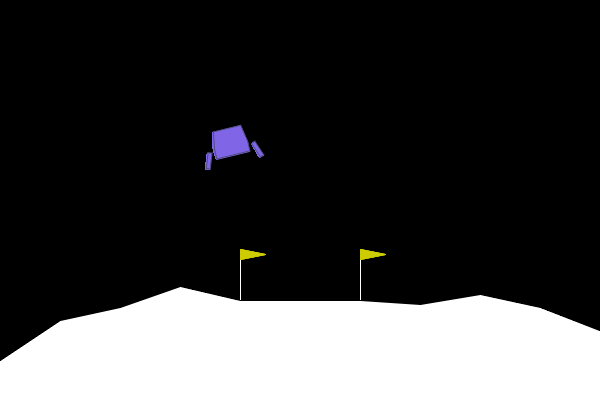}
        \end{subfigure} 
        \begin{subfigure}{1\textwidth}
            \centering
            \begin{tikzpicture}
                \begin{axis}[
                    grid=both,
                    width=\textwidth,
                    height=0.9\textwidth,
                    xtick=data,
                    xticklabels={
                        \(x\),\(y\),\(\dot{x}\),\(\dot{y}\),\(\theta\),\(\dot{\theta}\),$lc$,$rc$,$d_r$, $d_t$},
                    x tick label style={rotate=45, anchor=east},
                    enlarge x limits=0.05,
                    ybar,
                    bar width=10pt,
                    error bars/y dir=both,
                    error bars/y explicit
                ]
                \addplot[fill=red, draw=red] table {Plots/feature_viz/LunarLander-v2/RandomForestClassifier/info_2.dat};
                \end{axis}
            \end{tikzpicture}
        \end{subfigure}
    \end{subfigure}
    \caption{Feature importance scores coming from a trained RF behavior function computed for three different states of the \texttt{Lunar Lander} environment.}
        \label{fig:lunarlander_fi}
\end{figure*}

Finally, when it comes to the \texttt{Lunar Lander} environment we have observed that by far the most important feature of the environment is the one representing the $y$ position of the spaceship that is being controlled by the agent. This can be seen in all three histograms presented in Figure \ref{fig:lunarlander_fi}. Intuitively this is also a result that seems to make sense, as the goal of the agent is that of controlling the spaceship which is approaching the moon's surface from above. Interestingly, we can also note that the features modeling the left and right contact points of the spaceship, $l_c$ and $l_r$, have very low importance scores no matter how close the spaceship is to the surface. Furthermore, it can also be observed that when the spaceship is still far from approaching the moon's ground the importance scores of the features denoting the linear velocity $(\dot{y})$, as well as the angle $\theta$ and angular velocity $\dot{\theta}$ of the spaceship are fairly low. However, as highlighted by the third plot of Figure \ref{fig:lunarlander_fi} this changes once the spaceship is about to land. 

As the results presented in Figures \ref{fig:cartpole_fi}, \ref{fig:acrobot_fi} and \ref{fig:lunarlander_fi} are only a snapshot of the full agent-environment interaction, we refer the reader to the webpage of this project \footnote{\url{https://vimmoos-udrl.hf.space/}} where we release all trained behavior functions whose inference performance can be tested in the browser alongside the computation of the feature importance scores for all states encountered by the agent.

\section{\uppercase{Discussion \& Conclusion}}

In this paper, we have explored the potential that Upside Down Reinforcement Learning has to offer when it comes to the design of interpretable and transparent autonomous agents. We have shown that the behavior function $f$, which plays a crucial role in this novel learning paradigm, does not have to strictly come in the form of a neural network as originally described by \cite{schmidhuber2019reinforcement} and that tree-based ensemble methods, together with boosting algorithms, can perform just as well. We have also provided initial promising insights suggesting that behavior functions coming in the form of Random Forests and Extremely Randomized Trees can have much more to offer in terms of interpretability than their neural network counterparts. Based on these results we foresee several promising avenues for future work. The main, rather obvious, limitation of this paper is that it considers overall simple optimal-control benchmarks, which are far in terms of complexity from the RL environments that the DRL community adopts as test beds. How well UDRL agents trained with RFs and ETs would perform on high-dimensional and spatially organized state spaces such as those modeled by the popular Atari Arcade Learning Environment is unknown. While it is well known that convolutional neural networks have become the go-to design choice when it comes to these types of problems, it is worth mentioning that other valid alternatives could be explored. For example, in the contexts of forests of randomized trees, one could consider the work of \cite{maree2005random} and extend it from the value-based RL setting, where it has shown promising results on rather complex environments \citep{ernst2006reinforcement}, to the UDRL one. Next to scaling and generalizing the results presented in this study to more complex tasks, we believe another promising direction for future work revolves around coupling the tree-based behavior functions with other explanation tools. Techniques such as the recently introduced DPG \citep{arrighi2024decision} which helps with the identification of salient nodes in forests of randomized trees, or the more popular Shapley values \citep{winter2002shapley} alongside their extensions \citep{muschalik2024shapiq} come to mind. Lastly, it is worth highlighting the overall good performance obtained by the XGBoost algorithm. While we have focused our interpretability analysis on RFs and ETs, it is well-known that this algorithm also has a lot to offer regarding feature importance scores. In the future, we aim to provide a thorough comparison between different, inherently interpretable behavior functions, to further identify which state space components of an RL environment are the most important ones to a UDRL agent.

\clearpage
\twocolumn

\bibliographystyle{apalike}
{\small
\bibliography{example}}

\begin{thebibliography}{}

\bibitem[Agarwal and Mishra, 2021]{agarwal2021responsible}
Agarwal, S. and Mishra, S. (2021).
\newblock {\em Responsible AI}.
\newblock Springer.

\bibitem[Arrighi et~al., 2024]{arrighi2024decision}
Arrighi, L., Pennella, L., Marques~Tavares, G., and Barbon~Junior, S. (2024).
\newblock Decision predicate graphs: Enhancing interpretability in tree ensembles.
\newblock In {\em World Conference on Explainable Artificial Intelligence}, pages 311--332. Springer.

\bibitem[Arulkumaran et~al., 2022]{arulkumaran2022all}
Arulkumaran, K., Ashley, D.~R., Schmidhuber, J., and Srivastava, R.~K. (2022).
\newblock All you need is supervised learning: From imitation learning to meta-rl with upside down rl.
\newblock {\em arXiv preprint arXiv:2202.11960}.

\bibitem[Ashley et~al., 2022]{ashley2022learning}
Ashley, D.~R., Arulkumaran, K., Schmidhuber, J., and Srivastava, R.~K. (2022).
\newblock Learning relative return policies with upside-down reinforcement learning.
\newblock {\em arXiv preprint arXiv:2202.12742}.

\bibitem[Barto et~al., 1983]{barto1983neuronlike}
Barto, A.~G., Sutton, R.~S., and Anderson, C.~W. (1983).
\newblock Neuronlike adaptive elements that can solve difficult learning control problems.
\newblock {\em IEEE transactions on systems, man, and cybernetics}, (5):834--846.

\bibitem[Breiman, 2001]{breiman2001random}
Breiman, L. (2001).
\newblock Random forests.
\newblock {\em Machine learning}, 45:5--32.

\bibitem[Broniatowski et~al., 2021]{broniatowski2021psychological}
Broniatowski, D.~A. et~al. (2021).
\newblock Psychological foundations of explainability and interpretability in artificial intelligence.
\newblock {\em NIST, Tech. Rep}.

\bibitem[Brukhim et~al., 2022]{brukhim2022boosting}
Brukhim, N., Hazan, E., and Singh, K. (2022).
\newblock A boosting approach to reinforcement learning.
\newblock {\em Advances in Neural Information Processing Systems}, 35:33806--33817.

\bibitem[Busoniu et~al., 2017]{busoniu2017reinforcement}
Busoniu, L., Babuska, R., De~Schutter, B., and Ernst, D. (2017).
\newblock {\em Reinforcement learning and dynamic programming using function approximators}.
\newblock CRC press.

\bibitem[Chen et~al., 2021]{chen2021decision}
Chen, L., Lu, K., Rajeswaran, A., Lee, K., Grover, A., Laskin, M., Abbeel, P., Srinivas, A., and Mordatch, I. (2021).
\newblock Decision transformer: Reinforcement learning via sequence modeling.
\newblock {\em Advances in neural information processing systems}, 34:15084--15097.

\bibitem[Cover and Hart, 1967]{cover1967nearest}
Cover, T. and Hart, P. (1967).
\newblock Nearest neighbor pattern classification.
\newblock {\em IEEE transactions on information theory}, 13(1):21--27.

\bibitem[Degrave et~al., 2022]{degrave2022magnetic}
Degrave, J., Felici, F., Buchli, J., Neunert, M., Tracey, B., Carpanese, F., Ewalds, T., Hafner, R., Abdolmaleki, A., de~Las~Casas, D., et~al. (2022).
\newblock Magnetic control of tokamak plasmas through deep reinforcement learning.
\newblock {\em Nature}, 602(7897):414--419.

\bibitem[Ernst et~al., 2005]{ernst2005tree}
Ernst, D., Geurts, P., and Wehenkel, L. (2005).
\newblock Tree-based batch mode reinforcement learning.
\newblock {\em Journal of Machine Learning Research}, 6.

\bibitem[Ernst et~al., 2006a]{ernst2006reinforcement}
Ernst, D., Mar{\'e}e, R., and Wehenkel, L. (2006a).
\newblock Reinforcement learning with raw image pixels as input state.
\newblock In {\em Advances in Machine Vision, Image Processing, and Pattern Analysis: International Workshop on Intelligent Computing in Pattern Analysis/Synthesis, IWICPAS 2006 Xi’an, China, August 26-27, 2006 Proceedings}, pages 446--454. Springer.

\bibitem[Ernst et~al., 2006b]{ernst2006clinical}
Ernst, D., Stan, G.-B., Goncalves, J., and Wehenkel, L. (2006b).
\newblock Clinical data based optimal sti strategies for hiv: a reinforcement learning approach.
\newblock In {\em Proceedings of the 45th IEEE Conference on Decision and Control}, pages 667--672. IEEE.

\bibitem[Freund et~al., 1996]{freund1996experiments}
Freund, Y., Schapire, R.~E., et~al. (1996).
\newblock Experiments with a new boosting algorithm.
\newblock In {\em icml}, volume~96, pages 148--156. Citeseer.

\bibitem[Friedman, 2001]{friedman2001greedy}
Friedman, J.~H. (2001).
\newblock Greedy function approximation: a gradient boosting machine.
\newblock {\em Annals of statistics}, pages 1189--1232.

\bibitem[Furuta et~al., 2021]{furuta2021generalized}
Furuta, H., Matsuo, Y., and Gu, S.~S. (2021).
\newblock Generalized decision transformer for offline hindsight information matching.
\newblock {\em arXiv preprint arXiv:2111.10364}.

\bibitem[Geurts et~al., 2006]{geurts2006extremely}
Geurts, P., Ernst, D., and Wehenkel, L. (2006).
\newblock Extremely randomized trees.
\newblock {\em Machine learning}, 63:3--42.

\bibitem[Glanois et~al., 2024]{glanois2024survey}
Glanois, C., Weng, P., Zimmer, M., Li, D., Yang, T., Hao, J., and Liu, W. (2024).
\newblock A survey on interpretable reinforcement learning.
\newblock {\em Machine Learning}, pages 1--44.

\bibitem[HLEG, 2019]{eu_ethics_guidelines}
HLEG, H.-L. E. G. o.~A. (2019).
\newblock Ethics guidelines for trustworthy ai.

\bibitem[Ibrahim et~al., 2019]{ibrahim2019global}
Ibrahim, M., Louie, M., Modarres, C., and Paisley, J. (2019).
\newblock Global explanations of neural networks: Mapping the landscape of predictions.
\newblock In {\em Proceedings of the 2019 AAAI/ACM Conference on AI, Ethics, and Society}, pages 279--287.

\bibitem[James et~al., 2013]{james2013introduction}
James, G., Witten, D., Hastie, T., Tibshirani, R., et~al. (2013).
\newblock {\em An introduction to statistical learning}, volume 112.
\newblock Springer.

\bibitem[Leroy et~al., 2024]{leroy2024imp}
Leroy, P., Morato, P.~G., Pisane, J., Kolios, A., and Ernst, D. (2024).
\newblock Imp-marl: a suite of environments for large-scale infrastructure management planning via marl.
\newblock {\em Advances in Neural Information Processing Systems}, 36.

\bibitem[Liu et~al., 2022]{liu2022goal}
Liu, M., Zhu, M., and Zhang, W. (2022).
\newblock Goal-conditioned reinforcement learning: Problems and solutions.
\newblock {\em arXiv preprint arXiv:2201.08299}.

\bibitem[Louppe et~al., 2013]{louppe2013understanding}
Louppe, G., Wehenkel, L., Sutera, A., and Geurts, P. (2013).
\newblock Understanding variable importances in forests of randomized trees.
\newblock {\em Advances in neural information processing systems}, 26.

\bibitem[Mar{\'e}e et~al., 2005]{maree2005random}
Mar{\'e}e, R., Geurts, P., and Wehenkel, J. P.~L. (2005).
\newblock Random subwindows for robust image classification.
\newblock In {\em 2005 IEEE Computer Society Conference on Computer Vision and Pattern Recognition (CVPR'05)}, volume~1, pages 34--40. IEEE.

\bibitem[Mart{\'\i}n~H et~al., 2009]{martin2009k}
Mart{\'\i}n~H, J.~A., de~Lope, J., and Maravall, D. (2009).
\newblock The k nn-td reinforcement learning algorithm.
\newblock In {\em Methods and Models in Artificial and Natural Computation. A Homage to Professor Mira’s Scientific Legacy: Third International Work-Conference on the Interplay Between Natural and Artificial Computation, IWINAC 2009, Santiago de Compostela, Spain, June 22-26, 2009, Proceedings, Part I 3}, pages 305--314. Springer.

\bibitem[Min and Elliott, 2022]{min2022q}
Min, J. and Elliott, L.~T. (2022).
\newblock Q-learning with online random forests.
\newblock {\em arXiv preprint arXiv:2204.03771}.

\bibitem[Mirchevska et~al., 2017]{mirchevska2017reinforcement}
Mirchevska, B., Blum, M., Louis, L., Boedecker, J., and Werling, M. (2017).
\newblock Reinforcement learning for autonomous maneuvering in highway scenarios.
\newblock In {\em Workshop for Driving Assistance Systems and Autonomous Driving}, pages 32--41.

\bibitem[Miró-Nicolau et~al., 2025]{mironicolau2025comprehensive}
Miró-Nicolau, M., i~Capó, A.~J., and Moyà-Alcover, G. (2025).
\newblock A comprehensive study on fidelity metrics for xai.
\newblock {\em Information Processing \& Management}, 62(1):103900.

\bibitem[Muschalik et~al., 2024]{muschalik2024shapiq}
Muschalik, M., Baniecki, H., Fumagalli, F., Kolpaczki, P., Hammer, B., and H\"{u}llermeier, E. (2024).
\newblock shapiq: Shapley interactions for machine learning.
\newblock In {\em The Thirty-eight Conference on Neural Information Processing Systems Datasets and Benchmarks Track}.

\bibitem[Nauta et~al., 2023]{nauta2023anecdotal}
Nauta, M., Trienes, J., Pathak, S., Nguyen, E., Peters, M., Schmitt, Y., Schl{\"o}tterer, J., Van~Keulen, M., and Seifert, C. (2023).
\newblock From anecdotal evidence to quantitative evaluation methods: A systematic review on evaluating explainable ai.
\newblock {\em ACM Computing Surveys}, 55(13s):1--42.

\bibitem[Paszke et~al., 2017]{paszke2017automatic}
Paszke, A., Gross, S., Chintala, S., Chanan, G., Yang, E., DeVito, Z., Lin, Z., Desmaison, A., Antiga, L., and Lerer, A. (2017).
\newblock Automatic differentiation in pytorch.

\bibitem[Pedregosa et~al., 2011]{pedregosa2011scikit}
Pedregosa, F., Varoquaux, G., Gramfort, A., Michel, V., Thirion, B., Grisel, O., Blondel, M., Prettenhofer, P., Weiss, R., Dubourg, V., et~al. (2011).
\newblock Scikit-learn: Machine learning in python.
\newblock {\em the Journal of machine Learning research}, 12:2825--2830.

\bibitem[Prasad et~al., 2017]{prasad2017reinforcement}
Prasad, N., Cheng, L.-F., Chivers, C., Draugelis, M., and Engelhardt, B.~E. (2017).
\newblock A reinforcement learning approach to weaning of mechanical ventilation in intensive care units.
\newblock {\em arXiv preprint arXiv:1704.06300}.

\bibitem[Puterman, 2014]{puterman2014markov}
Puterman, M.~L. (2014).
\newblock {\em Markov decision processes: discrete stochastic dynamic programming}.
\newblock John Wiley \& Sons.

\bibitem[Rudin, 2019]{rudin2019stop}
Rudin, C. (2019).
\newblock Stop explaining black box machine learning models for high stakes decisions and use interpretable models instead.
\newblock {\em Nature machine intelligence}, 1(5):206--215.

\bibitem[Sallab et~al., 2017]{sallab2017deep}
Sallab, A.~E., Abdou, M., Perot, E., and Yogamani, S. (2017).
\newblock Deep reinforcement learning framework for autonomous driving.
\newblock {\em arXiv preprint arXiv:1704.02532}.

\bibitem[Schmidhuber, 2019]{schmidhuber2019reinforcement}
Schmidhuber, J. (2019).
\newblock Reinforcement learning upside down: Don't predict rewards--just map them to actions.
\newblock {\em arXiv preprint arXiv:1912.02875}.

\bibitem[Shah and Xie, 2018]{shah2018q}
Shah, D. and Xie, Q. (2018).
\newblock Q-learning with nearest neighbors.
\newblock {\em Advances in Neural Information Processing Systems}, 31.

\bibitem[Shwartz-Ziv and Armon, 2022]{shwartz2022tabular}
Shwartz-Ziv, R. and Armon, A. (2022).
\newblock Tabular data: Deep learning is not all you need.
\newblock {\em Information Fusion}, 81:84--90.

\bibitem[Song and Wang, 2024]{song2024multiobjective}
Song, Y. and Wang, L. (2024).
\newblock Multiobjective tree-based reinforcement learning for estimating tolerant dynamic treatment regimes.
\newblock {\em Biometrics}, 80(1):ujad017.

\bibitem[Srivastava et~al., 2019]{srivastava2019training}
Srivastava, R.~K., Shyam, P., Mutz, F., Ja{\'s}kowski, W., and Schmidhuber, J. (2019).
\newblock Training agents using upside-down reinforcement learning.
\newblock {\em arXiv preprint arXiv:1912.02877}.

\bibitem[Sutton, 1995]{sutton1995generalization}
Sutton, R.~S. (1995).
\newblock Generalization in reinforcement learning: Successful examples using sparse coarse coding.
\newblock {\em Advances in neural information processing systems}, 8.

\bibitem[Wehenkel et~al., 2006]{wehenkel2006ensembles}
Wehenkel, L., Ernst, D., and Geurts, P. (2006).
\newblock Ensembles of extremely randomized trees and some generic applications.
\newblock In {\em Robust methods for power system state estimation and load forecasting}.

\bibitem[Winter, 2002]{winter2002shapley}
Winter, E. (2002).
\newblock The shapley value.
\newblock {\em Handbook of game theory with economic applications}, 3:2025--2054.

\bibitem[Yu et~al., 2021]{yu2021reinforcement}
Yu, C., Liu, J., Nemati, S., and Yin, G. (2021).
\newblock Reinforcement learning in healthcare: A survey.
\newblock {\em ACM Computing Surveys (CSUR)}, 55(1):1--36.

\bibitem[Zhao et~al., 2009]{zhao2009reinforcement}
Zhao, Y., Kosorok, M.~R., and Zeng, D. (2009).
\newblock Reinforcement learning design for cancer clinical trials.
\newblock {\em Statistics in medicine}, 28(26):3294--3315.

\end{thebibliography}

\newpage
\onecolumn

\section*{\uppercase{Appendix}}

\begin{algorithm}[H]
\caption{Upside-Down Reinforcement Learning.}
\KwIn{$f_\theta$: Behavior function (e.g., neural network, random forest)} 
\KwIn{$\mathcal{M}$: Memory buffer with capacity $M$}
\KwIn{$\epsilon$: Exploration rate}

\SetKwFunction{UpdateBehaviorFunction}{UpdateBehaviorFunction}
\SetKwFunction{ResetEnv}{ResetEnvironment}
\SetKwFunction{GetBestEpisodes}{GetBestEpisodes}
\SetKwFunction{RandomInt}{RandomInt}
\SetKwFunction{Sum}{Sum}
\SetKwFunction{RandomAction}{RandomAction}
\SetKwFunction{MeanHorizon}{MeanHorizon}
\SetKwFunction{MeanReturn}{MeanReturn}
\SetKwFunction{StdReturn}{StdReturn}
\SetKwFunction{Step}{Step}
\SetKwFunction{Uniform}{\(\mathcal{U}\)}

\SetKwProg{Procedure}{Procedure}{}{}

\Procedure{Train}{episodes}{
    Initialize $\mathcal{S}, \mathcal{C}, \mathcal{A} \gets \emptyset, \emptyset, \emptyset$ \tcp*[r]{Training data}
    \ForEach{episode in episodes}{
        $T \gets$ Length(episode)\;
        $t_1 \gets \RandomInt(0, T-1)$\;
        $t_2 \gets T$\;
        $d_r \gets \Sum_{i=t_1}^{t_2} reward_i$\;
        $d_t \gets t_2 - t_1$\;
        $\mathcal{S} \gets \mathcal{S} \cup \{state_{t_1}\}$\;
        $\mathcal{C} \gets \mathcal{C} \cup \{[d_r, d_t]\}$\;
        $\mathcal{A} \gets \mathcal{A} \cup \{action_{t_1}\}$\;
    }
    \UpdateBehaviorFunction{$f_\theta, \mathcal{S}, \mathcal{C}, \mathcal{A}$}\;
}

\Procedure{CollectEpisode}{$d_r$, $d_t$}{
    $s_0 \gets \ResetEnv$\;
    $\mathcal{T} \gets \emptyset$ \tcp*[r]{Episode transitions}
    \While{not terminal}{
        $c \gets [d_r, d_t]$ \tcp*[r]{Command vector}
        \If{random or $\text{rand}() < \epsilon$}{
            $a \gets \RandomAction()$\;
        }
        \Else{
            $a \gets f_\theta(s, c)$\;
        }
        $s', r, \text{done} \gets \Step(a)$\;
        $\mathcal{T} \gets \mathcal{T} \cup \{(s, a, r)\}$\;
        $d_r \gets d_r - r$ \tcp*[r]{Update desired return}
        $d_t \gets \max(d_t - 1, 1)$ \tcp*[r]{Update desired horizon}
        $s \gets s'$\;
    }
    $\mathcal{M} \gets \mathcal{M} \cup \mathcal{T}$ \tcp*[r]{Store episode}
}

\Procedure{SampleCommands}{$\mathcal{M}$}{
    $\mathcal{B} \gets \GetBestEpisodes(\mathcal{M}, k)$ \tcp*[r]{Get k best episodes}
    $d_{h} \gets \MeanHorizon(\mathcal{B})$\;
    $\bar{R} \gets \MeanReturn(\mathcal{B})$\;
    $\sigma_R \gets \StdReturn(\mathcal{B})$\;
    $d_r \gets \Uniform(\bar{R}, \bar{R} + \sigma_R)$ \tcp*[r]{Sample desired return}
    \Return $d_r, d_t$\;
}
\end{algorithm}
\label{algo:udrl}

\end{document}